\documentclass[journal]{IEEEtran}
\usepackage[utf8]{inputenc} 
\usepackage[T1]{fontenc}    
\usepackage{hyperref}       
\usepackage{url}            
\usepackage{booktabs}       
\usepackage{amsfonts}       
\usepackage{nicefrac}       
\usepackage{microtype}      
\usepackage{csvsimple}
\usepackage{amsopn}
\usepackage{amsthm}
\usepackage{etex}
\usepackage{graphicx}
\usepackage{endnotes}
\usepackage{hyperref}
\usepackage{epsfig,psfrag}
\usepackage{pst-all}
\usepackage{amssymb,amsfonts,upref,cite,epsf,color,bm}
\usepackage{graphicx}
\usepackage{color}
\usepackage{amsmath}
\usepackage{graphicx}
\usepackage{calc}
\usepackage{booktabs}
\usepackage{tikz}
\usepackage{pgfplots}
\newcommand\defeq{:=}

\usepackage{algorithm}
\usepackage{algpseudocode}
\floatname{algorithm}{Algorithm}
\algnewcommand\algorithmicinput{\textbf{Input:}}
\algnewcommand\INPUT{\item[\algorithmicinput]}
\algnewcommand\algorithmicoutput{\textbf{Output:}}
\algnewcommand\OUTPUT{\item[\algorithmicoutput]}

\DeclareMathOperator*{\argmin}{arg\;min}

\newcommand\vect[1]{\mathbf #1}
\newcommand{\va}{\vect{a}}  

\newcommand{\vc}{\vect{c}}

\newcommand{\vv}{\vect{v}}  
\newcommand{\vw}{\vect{w}}
\newcommand{\vx}{\vect{x}}  
  
\newcommand{\vz}{\vect{z}}
\newcommand{\nodeidx}{i}

\newcommand{\mA}{\mathbf{A}}

\newcommand{\samplesize}{m}

\newcommand{\signalsize}{n}
\newcommand{\noise}{\varepsilon}

\newcommand{\incidence}{\vect{D}}
\newcommand{\featurelen}{p}

\newcommand{\graphsigs}{\mathcal{W}}

\newcommand{\edges}{\mathcal{E}}
\newcommand{\edgeidx}{e}
\newcommand{\cluster}{\mathcal{C}}
\newcommand{\nodes}{\mathcal{V}}
\newcommand{\graph}{\mathcal{G}}
\newcommand{\trainingset}{\mathcal{M}}
\newcommand{\samplingset}{\mathcal{M}}

\newcommand{\sigdim}{p}

\newcommand{\partition}{\mathcal{F}}

\newcommand{\compbound}{\overline{\partial \partition}}
\newcommand{\gsignal}{\mathbf{w}}
\newcommand{\gdual}{\mathbf{u}}
\newcommand{\hatgsignal}{\widehat{\gsignal}}
\newcommand{\numnodes}{n}
\newcommand{\numedges}{q}
\newcommand{\glabel}{y}
\newcommand{\gindex}[1][i]{^{(#1)}}
\newcommand{\gweight}{A}
\newcommand{\sigdimens}{p}
\newcommand{\gfeature}{\mathbf{x}}
\newcommand{\gvariable}{\mathbf{v}}

\newtheorem{theorem}{Theorem}
\newtheorem{definition}[theorem]{Definition}

\newcommand{\edge}[2]{\{#1,#2\}}
\usepackage{setspace}

\title{Localized Linear Regression in Networked Data}

\author{Alexander Jung and Nguyen Tran 
\thanks{Authors are with the Department of Computer Science, Aalto University, Finland; firstname.lastname(at)aalto.fi}
}
\begin{document}
	\maketitle
\begin{abstract}
The network Lasso (nLasso) has been proposed recently as an efficient learning algorithm for massive networked data sets 
(big data over networks). It extends the well-known least absolute shrinkage and selection operator (Lasso) from learning 
sparse (generalized) linear models to network models. Efficient implementations of the nLasso have been obtained using 
convex optimization methods lending to scalable message passing protocols. In this paper, we analyze the statistical properties 
of nLasso when applied to localized linear regression problems involving networked data. Our main result is a sufficient 
condition on the network structure and available label information such that nLasso accurately learns a localized linear regression 
model from a few labeled data points. We also provide an implementation of nLasso for localized linear regression by specializing 
a primal-dual method for solving the convex (non-smooth) nLasso problem.
\end{abstract}

\vspace*{-8mm}
\section{Introduction}
\label{sec_intro}

The data arising in many important application domains can be modeled efficiently using some network structure. 
Examples of such networked data are found in signal processing where signal samples can be arranged as a chain, 
in image processing with pixels arranged on a grid, in wireless sensor networks where measurements conform to sensor 
proximity \cite{shuman2013,MallatBook,OppenheimSchaferBuck1998,Chamon2018}. Organizing data using networks 
is also used in knowledge bases (graphs) whose items are linked by relations \cite{WikiData2014,Sadeghi2017}.  

In what follows, we will represent networked data using an undirected ``empirical graph''. The nodes of the empirical graph 
represent individual data points (e.g., one image out of an entire collection) which are connected by edges according to some 
notion of similarity. This similarity might be induced by domain knowledge (e.g., friendship relations in 
social networks) or via probabilistic models ( \cite{gmsIcassp2017,koller2009probabilistic}. 

Beside their network structure, data points are typically characterized by features and labels. The features of data points 
are quantities that can be measured or computed efficiently (in an automated fashion). In contrast, the labels of data points 
are costly to acquire, involving human expert labor. 

We consider regression problems within which data points are characterized by features and a numeric label (or target). The 
goal is to learn an accurate predictor which maps the features of a data point to a predicted label. The learning of the predictor 
is based on the availability of a few data points with known labels. Facing partially labeled data is common since the acquisition 
of reliable label information is often costly (involving human expert labor). 

Accurate learning is particularly challenging in the high-dimensional regime \cite{Gross2005,NetMedNat2010}. 
Here, a key obstacle is the lack of a sufficient amount of samples which can be 
considered i.i.d. Using a network structure allows then to borrow statistical strength from different ``groups'' of samples which 
are not exactly i.i.d., but still statistically similar to some extent. 

The learning of an accurate predictor from a small number of labeled data points is enabled by exploiting the tendency of 
well-connected data points to have similar statistical properties. Such a clustering assumption, which underlies most (semi-) supervised 
machine learning methods \cite{BishopBook,SemiSupervisedBook}, requires any reasonable predictor to be nearly constant 
over well-connected subsets (clusters) of data points. The clustering assumption motivates the network Lasso (nLasso) as a form 
of empirical risk minimization \cite{NetworkLasso}. 


{\bf Contribution.} While several implementations of nLasso have been proposed and analyzed (see \cite{NetworkLasso,pmlr-v54-yamada17a}), 
little is known about the accuracy of nLasso in regression problems. The main contribution of this paper is a sufficient condition on the network 
topology and available label information such that the nLasso accurately learns a predictor from a small number of labeled data points. 
To this end, we apply (an extension of) the network compatibility condition (NCC) introduced in \cite{WhenIsNLASSO}. 

We demonstrate theoretically and empirically, that the NCC guarantees that nLasso learns an accurate predictor 
which conforms with the clustering hypothesis. Our theoretical findings help to design sampling schemes 
which identify those data points whose labels would provide the most information about the labels of the other data 
points \cite{Marques2015,Chamon2018}.   

{\bf Notation.} 
The identity matrix of size $d\!\times\!d$ is denoted $\mathbf{I}_{d}$. The positive part of some real number $a\!\in\!\mathbb{R}$ is $(a)_{+} \!=\! \max\{a, 0\}$. The Euclidean 
norm of a vector $\vx\!=\!(x_{1},\ldots,x_{\featurelen})^{T}$ is $\| \vx \|\!\defeq\!\sqrt{\sum_{r=1}^{\featurelen} x_{r}^{2}}$. 
For a positive definite matrix $\mathbf{C}$, we define the induced norm $\| \vx \|_{\mathbf{C}} \defeq \sqrt{ \vx^{T} \mathbf{C} \vx }$. 
We will need the vector-valued clipping function $\mathcal{T}^{(\lambda)}(\vx) \defeq \lambda \vx/\|\vx\|$ for $\| \vx \| \geq \lambda$ and 
$\mathcal{T}^{(\lambda)}(\vx) \defeq \vx$ otherwise. The soft-thresholding operator is $\mathcal{S}(x;\tau)\defeq {\rm sign}(x) (|x|- \tau)_{+}$. 




\vspace*{-3mm}
\section{Problem Formulation}
\label{sec_setup}

We consider networked data modelled by an undirected ``empirical'' graph $\graph\!=\!(\nodes,\edges,\mA)$ 
whose nodes $\nodes\!=\!\{1,\ldots,\signalsize\}$ represent individual data points. The undirected edges 
$\edges$ encode some domain-specific notion of similarity between data points. The similarity between 
nodes $i,j\!\in\!\nodes$ connected by the edge $\{i,j\}\!\in\!\edges$ is quantified by a positive edge weight 
$A_{ij}$. We collect the weights (with $A_{ij}\!=\!0$ if nodes $i,j\!\in\!\nodes$ are not connected by an 
edge), into the weight matrix $\mA \in \mathbb{R}_{+}^{\signalsize \times \signalsize}$. 

In addition to the graph structure $\graph$, datasets typically convey additional information 
about the data points. Let us assume that each individual data point $i \in \nodes$ is 
characterized by a feature vectors $\vx^{(\nodeidx)} \in \mathbb{R}^{\featurelen}$ and a 
numeric label $y^{(\nodeidx)} \in \mathbb{R}$. The features $\vx^{(\nodeidx)}$ can be 
determined easily for any data point. In contrast, acquisition of labels $y^{(\nodeidx)}$ is 
difficult (requiring human expert labor). Our approach allows to have access 
only to the labels of a small training set $\trainingset=\{\nodeidx_{1},\ldots,\nodeidx_{\samplesize}\} \subseteq \nodes$.  

We relate features $\vx^{(\nodeidx)}$ and labels $y^{(\nodeidx)}$ using the linear model 
\vspace*{-2mm}
\begin{equation}
\label{equ_lin_model}
y^{(\nodeidx)} = \big( \overline{\vw}^{(\nodeidx)} \big)^{T} \vx^{(\nodeidx)}\!+\!\varepsilon^{(\nodeidx)},  
\end{equation} 
with some (unknown) weight vector $\overline{\vw}^{(\nodeidx)}$ for each node $\nodeidx \in \nodes$. 
The noise component $\varepsilon^{(\nodeidx)}$ in \eqref{equ_lin_model} summarizes any labeling 
our modeling errors. 

Thus, we assign each data point with an individual 
linear model \eqref{equ_lin_model}. For high-dimensional data (feature vector length $\featurelen$) this  
would result in overfitting unless we leverage the information contained in the network structure relating 
different data points. As we demonstrate theoretically and empirically, enforcing the (estimates of the) weight 
vectors $\overline{\vw}^{(\nodeidx)}$ to be similar for well-connected data points allows to accurately 
learn the linear models \eqref{equ_lin_model} for the entire dataset. 


We will apply nLasso to the available labels $y^{(\nodeidx)}$ for the training set to obtain an estimate $\widehat{\vw}^{(\nodeidx)}$ for the 
weight vector $\vw^{(\nodeidx)}$ at each node $\nodeidx \in \nodes$. The estimates $\widehat{\vw}^{(\nodeidx)}$ define a predictor which maps 
the node $\nodeidx \in \nodes$ to the predicted label 
\begin{equation}
\label{equ_predicted_label}
\hat{y}^{(\nodeidx)} \defeq \big( \widehat{\vw}^{(\nodeidx)} \big)^{T} \vx^{(\nodeidx)}. 
\end{equation}

The predictions $\hat{y}^{(\nodeidx)}$ will be accurate, i.e., the prediction error $\hat{y}^{(\nodeidx)} - y^{(\nodeidx)}$ will be small, 
if the estimation error $\overline{\vw}^{(\nodeidx)}\!-\!\widehat{\vw}^{(\nodeidx)}$ is small. Our main result (see Theorem \ref{lem_NSP1}) 
provides a sufficient condition on the structure of the empirical graph $\graph$ and the training set $\trainingset$ such that the 
estimation error is small. 

We interpret the weight vectors $\vw^{(\nodeidx)}$ as the values of a graph signal $\gsignal: \nodes \rightarrow \mathbb{R}^{\sigdimens}$ which 
assigns node $\nodeidx\!\in\!\nodes$ the vector $\gsignal\gindex\!\in\!\mathbb{R}^{\sigdimens}$. 
The set of all vector-valued graph signals is denoted 
\begin{equation} 
\label{equ_def_graph_sigs} 
\graphsigs\!\defeq\!\{ \gsignal: \nodes \rightarrow \mathbb{R}^{\sigdimens}: i \mapsto \gsignal \gindex \}.
\end{equation}  
Each graph signal $\widehat{\gsignal} \in \graphsigs$ represents a predictor which maps a node with features
 $\gfeature \gindex$ to the predicted label \eqref{equ_predicted_label}.

Given partially labeled networked data, we aim at leaning a predictor $\widehat{\gsignal} \in \graphsigs$ whose predictions \eqref{equ_predicted_label} agree 
with the labels $\glabel\gindex$ of labeled data points in the training set $\samplingset$. In particular, we aim at learning a predictor having a small training error 
\begin{align} 
\label{equ_def_emp_risk}
 \widehat{E}(\widehat{\gsignal})\!&\defeq\! \sum_{\nodeidx \in \samplingset} \big| y^{(\nodeidx)} - \hat{y}^{(\nodeidx)} \big| 
 \stackrel{\eqref{equ_predicted_label}}{=}  \sum_{\nodeidx \in \samplingset}  \big| y^{(\nodeidx)} - \big( \widehat{\vw}^{(\nodeidx)}\big)^{T} \vx^{(\nodeidx)} \big|. 
\end{align} 
We use the absolute value loss since it somewhat simplifies our analysis. However, we expect no big 
challenges in extending our analysis to nLasso using different loss functions, such as the squared 
error loss. The absolute value loss is actually preferred for learning linear regression models \eqref{equ_lin_model} 
when the noise $\noise^{(\nodeidx)}$ is expected to contain only a few large values, known as 
``salt and pepper'' noise in image processing \cite{pock_chambolle_2016}. 

\vspace*{-3mm}
\section{Network Lasso}
\vspace*{-1mm}

The criterion \eqref{equ_def_emp_risk} by itself is not enough for guiding the learning of a predictor 
$\widehat{\gsignal}$ since \eqref{equ_def_emp_risk} completely ignores the weights $\widehat{\gsignal}\gindex$ 
at unlabeled nodes $\nodeidx \in \nodes \setminus \samplingset$. Therefore, we need to impose some
additional structure on the predictor $\widehat{\gsignal}$. To this end, we require the predictor 
$\widehat{\gsignal}$ to conform with the \emph{cluster structure} of the empirical graph $\graph$ \cite{NewmannBook,Decelle2011}.  

The extend by which a predictor $\widehat{\gsignal}\!\in\!\graphsigs$ conforms with $\graph$ can be 
measured by the total variation (TV)
\vspace*{-1mm}
\begin{align}
\label{equ_def_TV_norm}
\| \gsignal \|_{\rm TV} & \defeq \sum_{\{i,j\}\in \edges} \gweight_{ij} \| \gsignal\gindex[j] - \gsignal\gindex \|. 
\vspace*{-1mm}
\end{align}
If the weights $\gsignal\gindex$ are approximately constant over well-connected subsets of nodes, the predictor 
$\gsignal\!\in\!\graphsigs$ has small TV $\|\widehat{\gsignal}\|_{\rm TV}$. The restriction of \eqref{equ_def_TV_norm} 
to a subset $\mathcal{S}\!\subseteq\!\edges$ of edges is denoted $\| \gsignal \|_{\mathcal{S}}\!\defeq\!\sum_{\{i,j\}\in \mathcal{S}} \gweight_{ij} \| \gsignal\gindex[j] - \gsignal\gindex \|$.

We are led naturally to learning a predictor $\widehat{\gsignal}$ via the  \emph{regularized empirical risk minimization} (ERM)
\vspace*{-2mm}
\begin{align} \label{optProb}
\hatgsignal & \in \argmin_{\gsignal \in \graphsigs} \widehat{E}(\gsignal)  + \lambda \| \gsignal \|_{\rm TV}, 
\end{align}
which is a special case of nLasso \cite{NetworkLasso}. 
The parameter $\lambda>0$ in \eqref{optProb} allows to trade small TV  $\| \hatgsignal \|_{\rm TV}$ against small error 
$\widehat{E}(\hatgsignal)$ \eqref{equ_def_emp_risk}. The choice of $\lambda$ can be guided by cross validation 
\cite{hastie01statisticallearning}. Alternatively the choice of $\lambda$ can be guided by our analysis of the 
nLasso estimation error (see discussion after Theorem \ref{lem_NSP1}). 
	
Note that nLasso \eqref{optProb} does not enforce the labels $\glabel\gindex$ themselves to be clustered. Instead, 
it requires the predictor $\widehat{\gsignal}$, which is used to obtain predictions \eqref{equ_predicted_label}, to be 
clustered. 
	
It will be convenient to reformulate \eqref{optProb} using vector notation. To this end, we represent a graph signal $\gsignal \in \graphsigs$ 
as the vector
\begin{align}
\label{equ_def_vector_signal}
\gsignal = ((\gsignal\gindex[1])^T, \ldots ,(\gsignal\gindex[\numnodes])^T)^T \in \mathbb{R}^{\featurelen\numnodes}.
\end{align}
and define the block matrix $\incidence\!\in\!\mathbb{R}^{\featurelen \numedges \times \featurelen \numnodes}$ (with $\numedges\!=\!|\edges|$)
\begin{align}
\incidence_{e,i} = \begin{cases}
\gweight_{ij} \mathbf{I}_{\sigdimens} & e=\{i,j\}\in\edges, i<j\\
-\gweight_{ij} \mathbf{I}_{\sigdimens}& e=\{i,j\}\in\edges, i>j\\
\mathbf{0} & {\rm otherwise}.
\end{cases}
\label{equ_def_incident_matrix}
\end{align}
Applying the matrix $\mathbf{D}$ to a graph signal vector $\gsignal$ \eqref{equ_def_vector_signal} 
results in a partitioned vector $\mathbf{D} \gsignal$ whose $e$th block is given by $\gweight_{ij} (\gsignal\gindex - \gsignal\gindex[j])$  (see \eqref{equ_def_TV_norm}). 
Using \eqref{equ_def_vector_signal} and \eqref{equ_def_incident_matrix}, we can 
reformulate the nLasso \eqref{optProb} as
\begin{align}\label{LNLprob}
\hatgsignal \in \argmin_{\gsignal \in \mathbb{R}^{\sigdimens\numnodes}}   h(\gsignal) + g(\incidence \gsignal). 
\end{align}
Here, 
\vspace*{-4mm}
\begin{align}
\label{equ_def_opt_func}
h(\gsignal) & = \widehat{E}(\gsignal) \text{ and }  g(\gdual) \defeq \lambda \sum_{e=1}^{\numedges}\|\gdual^{(e)}\|  \\ 
&\mbox{ with } \gdual=\!\big( \big(\gdual^{(1)}\big)^{T},\ldots,\big(\gdual^{(\numedges)}\big)^{T} \big)^{T} \in \mathbb{R}^{\sigdimens\numedges}. \nonumber
\end{align}

	
\vspace*{-4mm}
\section{Primal-Dual Method}
\label{sec_lNLasso_ADMM}

The nLasso \eqref{LNLprob} is a convex optimization problem with a non-smooth objective 
function which rules out the use of gradient descent methods \cite{JungFixedPoint}. However, 
the objective function is highly structured since it is the sum of two components $h(\gsignal)$ 
and $g(\incidence \gsignal)$, which can be optimized efficiently when considered separately. 
Such composite functions can be optimized efficiently using proximal splitting methods \cite{Combettes2009, Connor2014, pock_chambolle}.
 
We apply the proximal method proposed in \cite{PrecPockChambolle2011} 
which is based on reformulating \eqref{LNLprob} as a saddle-point problem  
\begin{align}
\label{equ_pd_prob}
\min_{\gsignal \in \mathbb{R}^{\sigdimens \numnodes}} \max_{\gdual \in \mathbb{R}^{\sigdimens\numedges}} \gdual^T\incidence \gsignal  + h(\gsignal) - g^*(\gdual),
\end{align}
with the convex conjugate  $g^*$ of $g$ \cite{pock_chambolle}.

Solutions $(\hatgsignal, \widehat{\gdual})$ of \eqref{equ_pd_prob} are characterized by \cite[Thm 31.3]{RockafellarBook} 
\begin{align}
-\incidence^T \widehat{\gdual}  \in \partial h(\hatgsignal)\mbox{, and }\incidence \hatgsignal \in \partial g^*(\widehat{\gdual}).
\label{equ_pd_cond_1}
\end{align}
The coupled conditions \eqref{equ_pd_cond_1} are, in turn, equivalent to
\begin{align} 
\hspace*{-1.9mm}\hatgsignal\!-\!\mathbf{T} \incidence^T \widehat{\gdual}\!\in\!(\mathbf{I}\!+\!\mathbf{T} \partial h) (\hatgsignal)\mbox{, }
\widehat{\gdual}\!+\!\boldsymbol{\Sigma} \incidence \hatgsignal\!\in\!(\mathbf{I}\!+\!\boldsymbol{\Sigma} \partial g^*)(\widehat{\gdual}),
\label{equ_pd_cond_1_2}
\end{align}
with positive definite matrices $ \boldsymbol{\Sigma} \!\in\! \mathbb{R}^{\sigdimens \numedges \times \sigdimens\numedges}, \mathbf{T} \!\in\! \mathbb{R}^{\sigdimens\numnodes \times \sigdimens\numnodes}$. 
In principle, the matrices $\boldsymbol{\Sigma}, \mathbf{T}$ in \eqref{equ_pd_cond_1_2} can be chosen arbitrarily. 
It will prove convenient to choose them as
\begin{equation}
\label{equ_def_diag_matrix}
\boldsymbol{\Sigma} \!=\! {\rm diag} \{\sigma^{(\edgeidx)}\mathbf{I}_{\sigdimens}\}_{\edgeidx=1}^{\numedges} \mbox{ and } \mathbf{T}\!=\! {\rm diag} \{\tau^{(\nodeidx)}\mathbf{I}_{\sigdimens}\}_{\nodeidx=1}^{\numnodes}
\end{equation} 
with scalars $\big\{ \sigma^{(\edgeidx)} \big\}_{\edgeidx\!=\!1}^{\numedges}$ and $\big\{ \tau^{(\nodeidx)} \big\}_{\nodeidx \in \nodes}$ as specified below. 

The optimality condition \eqref{equ_pd_cond_1_2} for nLasso \eqref{LNLprob} 
lends naturally to the following coupled fixed point iterations \cite{PrecPockChambolle2011}
\begin{align}
\gsignal_{k+1} \!&=\! (\mathbf{I} \!+\! \mathbf{T} \partial h)^{-1} (\gsignal_{k} \!-\!\mathbf{T} \incidence^T \gdual_{k})  \label{equ_pd_upd_x} \\
\gdual_{k+1} \!&=\! (\mathbf{I} \!+\! \boldsymbol{\Sigma} \partial g^*)^{-1} (\gdual_{k} \!+\! \boldsymbol{\Sigma} \incidence (2\gsignal_{k+1} \!-\! \gsignal_{k})).
\label{equ_pd_upd_y}
\end{align}

The update \eqref{equ_pd_upd_y} involves the resolvent operator 
\begin{align}
\label{equ_def_prox}
\hspace*{-2mm}(\mathbf{I} \!+\! \boldsymbol{\Sigma} \partial g^*)^{-1} (\gdual) \!=\! \argmin_{\gdual' \in \mathbb{R}^{\sigdimens \numedges} } g^*(\gdual') \!+\! (1/2)\| \gdual' \!-\! \gdual\|^2_{\boldsymbol{\Sigma}^{-1}}.
\end{align}
The convex conjugate $g^*$ of $g$ (see \eqref{equ_def_opt_func}) can be decomposed 
as $g^*(\gvariable)\!=\!\sum\limits_{e\!=\!1}^{\numedges} g_2^*(\gvariable^{(e)})$ with the 
convex conjugate $g_2^*$ of $g_{2}(\vz) \defeq \lambda \|\vz\|$. Combining the fact that 
$\boldsymbol{\Sigma}$ is a block diagonal matrix with the Moreau decomposition \cite[Sec. 6.5]{ProximalMethods}, 
it can be shown that $\vc = (\mathbf{I}_{\sigdimens\numedges}\!+\!\boldsymbol{\Sigma} \partial g^*)^{-1} (\gdual)$ (see \eqref{equ_def_prox})
with  
\begin{align}
\label{equ_update_c_explicit}
\vc = \big( \big( \vc^{(1)}\big)^{T},\ldots,\big(\vc^{(\numedges)} \big)^{T} \big)^{T}\mbox{, } \mathbf{c}^{(e)} \defeq  \mathcal{T}^{(\lambda)}\big(\gdual^{(e)}\big). 
 \end{align}
 
Similar to the update \eqref{equ_pd_upd_y}, also the update \eqref{equ_pd_upd_x} decomposes into 
independent updates of the weight vectors  
\begin{equation}
\gsignal^{(\nodeidx)}\!=\!\gsignal^{(\nodeidx)}_{k} \!-\!\sum_{j > i} \tau^{(j)} A_{i,j} \gdual^{(j)}_{k} \!+\!\sum_{i > j} \tau^{(j)} A_{i,j} \gdual^{(j)}_{k} \nonumber
\end{equation}
yielding the updated weight vectors $\gsignal^{(\nodeidx)}_{k+1}  = \vv^{(\nodeidx)}$ for each node $\nodeidx \in \nodes$. 
In particular, for unlabeled nodes $\nodeidx \notin \trainingset$, the update \eqref{equ_pd_upd_x} reduces to 
$\vv^{(\nodeidx)}=\gsignal^{(\nodeidx)}$. For labeled nodes $\nodeidx \in \trainingset$, using elementary sub-gradient calculus, 
we obtain 
\begin{align}
\vv^{(\nodeidx)} & = \mathbf{x}^{(\nodeidx)}(\tilde{y}\!+\!\mathcal{S}(\tilde{w}\!-\!\tilde{y};\tau^{(\nodeidx)})) \nonumber \\
& + (\mathbf{I}\!-\!(1/\|\vx^{(\nodeidx)}\|^{2}) \vx^{(\nodeidx)}\big(\vx^{(\nodeidx)}\big)^{T})\gsignal^{(\nodeidx)}
\label{equ_closed_form_update_block_threshold}
\end{align}
with $\tilde{y}\defeq y^{(\nodeidx)}/\|\vx^{(\nodeidx)}\|^{2} $ and $\tilde{w} \defeq \big(\gsignal^{(\nodeidx)}\big)^{T} \vx^{(\nodeidx)} / \|\vx^{(\nodeidx)}\|^{2}$. 
Inserting \eqref{equ_closed_form_update_block_threshold} and \eqref{equ_update_c_explicit} into the 
fixed point iteration \eqref{equ_pd_upd_x}, \eqref{equ_pd_upd_y} results in Alg.\ \ref{alg:primal_dual} for solving the nLasso \eqref{LNLprob}. 

If the matrices $\boldsymbol{\Sigma}$ and $\mathbf{T}$ using in \eqref{equ_pd_upd_y} satisfy
\begin{align}
\|\boldsymbol{\Sigma}^{1/2} \incidence  \mathbf{T}^{1/2}\|^2 <1,
\label{equ_pre_cond}
\end{align}
the sequences obtained from iterating \eqref{equ_pd_upd_x} and \eqref{equ_pd_upd_y} converge to a saddle point of 
the problem \eqref{equ_pd_prob} \cite[Thm. 1]{PrecPockChambolle2011}. 
The condition \eqref{equ_pre_cond} is ensured by choosing $\boldsymbol{\Sigma}$ and $\mathbf{T}$ 
according to \eqref{equ_def_diag_matrix} using $\sigma^{(\edgeidx)}\!=\!1/(2\gweight_{\edgeidx})$ and $\tau^{(\nodeidx)}\!\defeq\!\eta / d^{(\nodeidx)}$, 
with (weighted) node degree $d^{(\nodeidx)}\!=\!\sum_{j\!\neq\!\nodeidx} \gweight_{\nodeidx,j}$ and some constant $\eta\!<\!1$ \cite[Lem. 2]{PrecPockChambolle2011}. 

\begin{algorithm}[]
\caption{nLasso via primal-dual method}\label{alg:primal_dual}
\begin{algorithmic}[1]
\renewcommand{\algorithmicrequire}{\textbf{Input:}}
\renewcommand{\algorithmicensure}{\textbf{Output:}}
\Require   $\graph = (\nodes, \edges, \mathbf{A})$, $\{\gfeature\gindex\}_{\nodeidx \in \nodes}$, $\samplingset$, 
$\{ \glabel\gindex \}_{i \in \samplingset}$, $\lambda$
\Statex\hspace{-6mm}{\bf Initialize:} $k\!\defeq\!0$, $\widehat{\gsignal}_0\!\defeq\!0$, $\widehat{\gdual}_0\!\defeq\!0$; $\boldsymbol{\Sigma}$ 
and $\mathbf{T}$ using \eqref{equ_def_diag_matrix} with $\sigma^{(e)}=1/(2\gweight_{e})$, $\tau^{(\nodeidx)} = 0.9/d^{(\nodeidx)}$; 
$\beta_{\nodeidx} \defeq \tau^{(\nodeidx)} /|\samplingset|$; incidence matrix $\incidence$ according to \eqref{equ_def_incident_matrix}

\Repeat
  \State \hspace*{-3mm} $\widehat{\gsignal}_{k+1} \defeq \widehat{\gsignal}_{k} - \mathbf{T} \incidence^T \widehat{\gdual}_{k}$
  \State \hspace*{-3mm} for each labeled node $\nodeidx\!\in\!\samplingset$ set $\widehat{\gsignal}_{k+1}\gindex\!\defeq\!\vv^{(\nodeidx)}$ using \eqref{equ_closed_form_update_block_threshold}  
   \State  \hspace*{-3mm} $\overline{\gdual} \defeq {\gdual}_k + \boldsymbol{\Sigma} \incidence (2\widehat{\gsignal}_{k+1}-\widehat{\gsignal}_{k})$ 

   \State  \hspace*{-3mm} for each edge $e\!\in\!\edges$ set $\widehat{\gdual}_{k+1}^{(e)}\!\defeq\!\mathcal{T}^{(\lambda)} \big( \overline{\gdual}^{(e)} \big)$  

  \State  \hspace*{-3mm}  $k\!\defeq\!k\!+\!1$
   \Until stopping criterion is satisfied 
   \Ensure predictor $\widehat{\vw} \defeq \hatgsignal_{k}$
\end{algorithmic}
\end{algorithm}

Another instance of a proximal method is the alternating direction method of multipliers (ADMM) \cite{ProximalMethods,DistrOptStatistLearningADMM}, 
which has been applied to (a more general formulation of) the nLasso in \cite{NetworkLasso}. In 
contrast, to the primal-dual method used in Alg.\ \ref{alg:primal_dual}, the ADMM implementation 
involves a tuning parameter. The optimum choice for this tuning parameter is non-trivial and typically 
requires a grid search \cite{Nishihara2015}. However, we expect that Alg.\ \ref{alg:primal_dual} and 
the ADMM implementation of \cite{NetworkLasso} (when specialized to \eqref{equ_def_emp_risk}) 
to have similar computational requirements.

\vspace*{-3mm}
\section{Error Analysis for nLasso} 
\vspace*{-1mm}

In order to analyze the statistical properties of Alg.\ \ref{alg:primal_dual} we need to understand the 
structure of the solutions to the nLasso problem \eqref{LNLprob}. To this end, will use a simple but useful 
model of piece-wise constant weight vectors     
\begin{equation}
\label{equ_def_clustered_signal_model}
 \overline{\vw}^{(\nodeidx)} \!=\! \sum_{l=1}^{F} \va^{(l)} \mathcal{I}_{\cluster^{(l)}}[\nodeidx]. 
\end{equation}  
with fixed vectors $\va^{(l)} \in \mathbb{R}^{\sigdim}$, for $l=1,\ldots,F$, and the indicator function $\mathcal{I}_{\cluster}[\nodeidx] \in \{0,1\}$ with 
$\mathcal{I}_{\cluster}[\nodeidx] = 1$ if and only if $\nodeidx \in \cluster \subseteq \nodes$. 
Here, we use a partition $\partition = \{\cluster^{(1)},\ldots,\cluster^{(F)}\}$ of the nodes $\nodes$ in the empirical graph 
into disjoint subsets (clusters) $\cluster^{(l)}$. 

The model \eqref{equ_def_clustered_signal_model}, which generalizes the piece-wise constant signal model 
(see \cite{FanGuan2017,ChenClustered2016}), embodies a clustering assumption that well-connected nodes 
in the empirical graph should have similar relations between features and labels \cite{Decelle2011,NewmannBook}. 

Note that our analysis allows for an arbitrary choice of clusters $\cluster^{(l)}$ in \eqref{equ_def_clustered_signal_model}. 
However, our results are most useful when the sets $\cluster^{(l)}$ reflect the intrinsic cluster structure of the empirical 
graph $\graph$ such that the TV $\| \overline{\vw} \|_{\rm TV}$ (see \eqref{equ_def_TV_norm}) is small.

We now introduce the network compatibility condition (NCC), which generalizes the 
compatibility conditions for Lasso type estimators \cite{BuhlGeerBook} of ordinary sparse signals. 
Our main contribution is to show that the NCC guarantees the accuracy of the nLasso \eqref{LNLprob} 
solutions, as obtained using Alg.\ \ref{alg:primal_dual}. 

\begin{definition} 
\label{def_NNSP}
Consider a networked dataset with empirical graph $\graph = (\nodes, \edges,\mA)$. The nodes are characterized by 
feature vectors $\vx^{(\nodeidx)} \in \mathbb{R}^{\featurelen}$ and grouped according to a fixed partition $\partition=\{\cluster^{(1)},\ldots,\cluster^{(F)}\}$. 
The labels $y^{(\nodeidx)}$ of nodes are observed only on the training set $\samplingset \subseteq \nodes$. 
The training set is said to satisfy NCC, with constants $K,L>0$, if 
\vspace*{-1mm}
\begin{equation}
\label{equ_ineq_multcompcondition_condition}
K \sum_{\nodeidx \in \samplingset} \big| \big( \vx^{(\nodeidx)}\big)^{T} \vw^{(\nodeidx)}\big|  +  \| \vw \|_{\compbound} \geq (L/\sqrt{\featurelen})  \| \vw \|_{\partial \partition} 
\vspace*{-1mm}
\end{equation} 
for any graph signal $\vw \in \graphsigs$ (see \eqref{equ_def_graph_sigs}). 
\end{definition} 
We highlight that the NCC (constants) depend jointly on the training set $\samplingset$ 
and the  network structure of $\graph$. While enlarging the training set can only improve 
the NCC constants (smaller $K$), the precise quantification of this improvement is difficult.
 
As shown in \cite{WhenIsNLASSO,NNSPSampta2017}, the NCC is satisfied if there exists a sufficiently 
large network flow between sampled nodes. Thus, given a dataset with empirical graph $\graph$, the NCC 
can be verified using network flow algorithms (see Section \ref{sec_numexp} and \cite{GoldbergTarjan2014}). 

Our main theoretical result is that if the sampling set satisfies the NCC (see Definition \ref{def_NNSP}), any solution of 
\eqref{optProb} is close to the true underlying weight vectors (see \eqref{equ_lin_model}, \eqref{equ_def_clustered_signal_model}). 
\vspace*{-2mm}
\begin{theorem} 
\label{lem_NSP1}
Consider a partially labeled networked dataset with empirical graph $\graph$ with features $\vx^{(\nodeidx)}$ known for all nodes 
and labels $y^{(\nodeidx)}$ which are known only for the nodes $\nodeidx \in \trainingset$. We assume a linear model \eqref{equ_lin_model} 
with true weights $\overline{\vw}^{(\nodeidx)}$ piece-wise constant \eqref{equ_def_clustered_signal_model}. 
If the sampling set $\samplingset$ satisfies NCC with parameters $L > \sqrt{\sigdim}$ and $K > 0$, then any solution $\widehat{\vw}$ 
of nLasso \eqref{LNLprob} with the choice $\lambda \defeq 1/K$ satisfies  
\begin{equation}
 \| \widehat{\vw}-\overline{\vw} \|_{\rm TV} \!\leq\! K (1\!+\!4\sqrt{\sigdim}/(L\!-\!\sqrt{\sigdim}))  \sum_{\nodeidx \in \samplingset} |\noise^{(\nodeidx)}| .
 \label{equ_bound_error_TV}
\end{equation} 
\end{theorem}
According to Theorem \ref{lem_NSP1}, the choice for the nLasso parameter $\lambda$ in \eqref{LNLprob} can be based 
on the NCC constant $K$ (see \eqref{equ_ineq_multcompcondition_condition}) via setting $\lambda\!=\!1/K$. For this 
choice, given the training set $\samplingset$ satisfies the NCC with parameters $K$ and $L$, the nLasso error 
$\widehat{\vw}\!-\!\overline{\vw}$ is bounded according to \eqref{equ_bound_error_TV}. 

Note that the bound \eqref{equ_bound_error_TV} does neither explicitly involve the size $\samplesize\!=\!|\samplingset|$ 
of the training set $\samplingset$, nor the overall size $\signalsize$ of the empirical graph (or dataset). However, the relative 
size $m/\signalsize$ of the training set will influence the probability that the NCC is satisfied (such that the bound 
\eqref{equ_bound_error_TV} applies at all). 

We highlight that the nLasso \eqref{optProb} does not require the partition $\partition$ used for our signal model 
\eqref{equ_def_clustered_signal_model}. This partition is only used for the analysis of nLasso \eqref{optProb}. 
Moreover, if the true underlying graph signal is of the form \eqref{equ_def_clustered_signal_model} and nLasso 
accurately learns this signal, we can obtain the partition $\partition$ by thresholding the edge-wise differences 
$\| \vw^{(\nodeidx)} \!-\! \vw^{(j)} \|$ for $\edge{i}{j}\!\in\!\edges$ \cite{TrendGraph}. 

\vspace*{-4mm}
\section{Numerical Experiments} 
\label{sec_numexp}
In order to verify our theoretical findings (see Theorem \ref{lem_NSP1}), we have applied Alg.\ \ref{alg:primal_dual} to two particular 
datasets. The first dataset is synthetically generated based on an empirical graph which consists of two well-connected 
clusters. We also consider a dataset obtained from temperature 
measurements at various locations in Finland.\footnote{The source code for our numerical experiments can be found under \url{https://github.com/alexjungaalto/ResearchPublic/tree/master/LocalizedLinReg}.}

{\bf Two-Cluster Dataset.} We generate the empirical graph $\graph$ ($\signalsize\!=\!80$) by sparsely connecting two random graphs $\cluster^{(1)}$ 
and $\cluster^{(2)}$, each of size $\signalsize/2$ and with average degree $10$. 
The nodes of $\graph$ are assigned feature vectors $\vx^{(\nodeidx)} \in \mathbb{R}^{2}$ 
obtained by i.i.d.\ random vectors uniformly distributed on the unit sphere $\{ \vx \in \mathbb{R}^{2}: \| \vx \|=1\}$. 
The labels $y^{(\nodeidx)}$ of the nodes $i\in \nodes$ are generated according to the linear model \eqref{equ_lin_model} 
with zero noise $\varepsilon^{(\nodeidx)}=0$ and piecewise constant weight vectors $\mathbf{w}^{(\nodeidx)}$ 
(see \eqref{equ_def_clustered_signal_model}). We assume that the labels $y^{(\nodeidx)}$ are known for the nodes in 
the training set which includes three data points from each cluster, i.e., $|\samplingset \cap \cluster^{(1)}|= |\samplingset \cap \cluster^{(2)}|=3$. 

Using \cite[Lemma 6]{WhenIsNLASSO} it can be shown that the training set $\samplingset$ satisfies NCC with 
$L\!>\!\sqrt{\featurelen}\!=\!\sqrt{2}$ if there exists a sufficiently large network flow between the labeled node 
$i\!\in\!\cluster^{(l)}\!\cap\!\trainingset$ and the boundary edges $\partial \defeq \{ \{i,j\} \in \edges: i \in \cluster^{(1)}, j \in \cluster^{(2)} \}$ 
between the two clusters. In particular, let $\rho^{(l)}$ denote the normalized flow value from the labeled nodes 
in cluster $\cluster^{(l)}$ and the cluster boundary, normalized by the boundary size $|\partial|$. The NCC is satisfied 
with $L\!>\!\sqrt{2}$ if $\rho^{(l)}\!>\!\sqrt{2}$ for $l\!=\!1,2$.

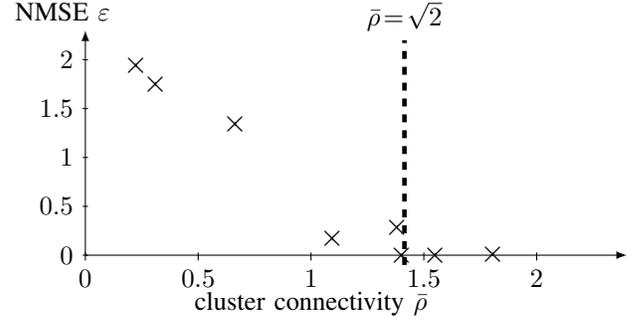
\begin{figure}[htbp]
\begin{center}
\begin{tikzpicture}
 \tikzset{x=3cm,y=1.3cm,every path/.style={>=latex},node style/.style={circle,draw}}
    \csvreader[ head to column names,%
                late after head=\xdef\aold{\a}\xdef\bold{\b},,%
                after line=\xdef\aold{\a}\xdef\bold{\b}]%
              {MSEoverBoundary_11-May-2019.csv}{}        
                {\draw [line width=0.5mm] (\a, \b) -- (\a,\b) node {\large $\times$};
    }
          \draw[->] (0,0) -- (2.4,0);
      \node [] at (1,-0.5) {\centering cluster connectivity $\bar{\rho}$};
      \draw[->] (0,0) -- (0,2.3);
      \node [anchor=south] at (-0.1,2.3) {NMSE $\varepsilon$};
            \foreach \label/\labelval in {0/$0$,0.5/$0.5$,1/$1$,1.5/$1.5$,2/$2$}
        { 
          \draw (1pt,\label) -- (-1pt,\label) node[left] {\labelval};
        }
        \foreach \label/\labelval in {0/$0$,0.5/$0.5$,1/$1$,1.5/$1.5$,2/$2$}
        { 
          \draw (\label,1pt) -- (\label,-2pt) node[below] {\labelval};
        }
        
        \draw[dashed, line width=.6mm] (1.414,-0.1) -- (1.414,2.2) ;
        \node[anchor=south] at (1.414,2.2) {$\bar{\rho}\!=\!\sqrt{2}$} ; 
\end{tikzpicture}
\end{center}
\vspace*{-4mm}
  \caption{NMSE achieved by Alg.\ \ref{alg:primal_dual} for a two-cluster graph. } 
  \label{fig_NMSEconnect}
  \vspace*{-7mm}
\end{figure}

In Fig.\ \ref{fig_NMSEconnect}, we depict the normalized mean squared error (NMSE) $\varepsilon\!\defeq\!\| \overline{\vw}\!-\!\widehat{\vw} \|^{2}_{2} / \| \overline{\vw} \|^{2}_{2}$ 
incurred by Alg.\ \ref{alg:primal_dual} (averaged over $10$ i.i.d.\ simulation runs) for varying connectivity, as 
measured by the empirical average $\bar{\rho}$ of $\rho^{(1)}$ and $\rho^{(2)}$ (having same distribution). Note that 
Fig.\ \ref{fig_NMSEconnect} agrees with Theorem \ref{lem_NSP1} which predicts Alg.\ \ref{alg:primal_dual} is accurate if NCC holds ($\bar{\rho}\!>\!\sqrt{2}$ ). 

\vspace*{-4mm}

\begin{figure}[htbp]
\begin{center}
 \includegraphics[height=3.8cm,width=6cm]{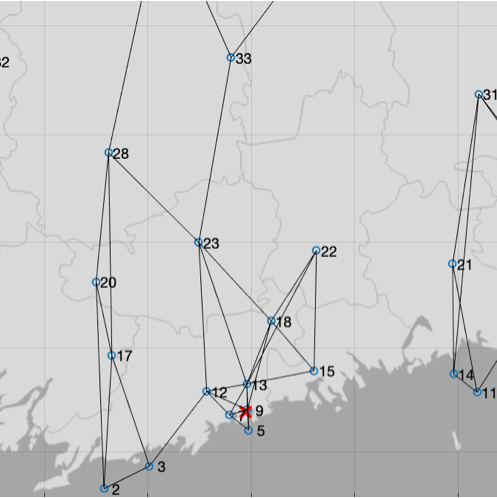}  
 \vspace*{-2mm}
\caption{Weather stations in Finland.}
\label{fig:FMI}
\end{center}
\vspace*{-4mm}
\end{figure}
{\bf Weather Data.} In this experiment, we consider a networked dataset whose empirical graph $\graph$ represents 
Finnish weather stations (see Fig.\ \ref{fig:FMI}), which are initially connected by an edge to their $K=3$ nearest 
neighbors. The feature vector $\vx^{(i)}\!\in\!\mathbb{R}^{3}$ of node $i\!\in\!\nodes$ contains the local (daily mean) 
temperature for the preceding three days. The label $y^{(i)} \in \mathbb{R}$ is the current day-average temperature. 

We use Alg.\ \ref{alg:primal_dual} to learn the weight vectors $\vw^{(i)}$ for a localized linear model \eqref{equ_lin_model}. 
For the sake of illustration we focus on the weather stations in the capital region around Helsinki (indicated by a red cross in Fig.\ \ref{fig:FMI}). 
These stations are represented by nodes $\cluster\!=\!\{23,18,22,15,12,13,9,7,5\}$ and we assume that labels $y^{(i)}$ are available for all 
nodes outside $\cluster$ and for the nodes $i\!\in\! \{12,13,15\}\!\subseteq\!\cluster$. Thus, for more than half of the nodes in $\cluster$ we do 
not know the labels $y^{(i)}$ but predict them via \eqref{equ_predicted_label} with the weight vectors $\widehat{\vw}^{(i)}$ 
obtained from Alg.\ \ref{alg:primal_dual} (using $\lambda\!=1/7$ and a fixed number of $10^{4}$ iterations). The normalized 
average squared prediction error is  $\approx 10^{-1}$ and only slightly larger than the prediction error incurred by fitting 
a single linear model to the cluster $\cluster$ using a least absolute deviation regression method \cite[Sec. 6.1]{DistrOptStatistLearningADMM}. 

\section*{Acknowledgments}
We thank Roope Tervo from the Finnish Meteorological Institute for helping with gathering the weather data.

\bibliographystyle{IEEEtran}
\bibliography{LitLink}

\newpage
\vspace*{0mm}
\section{Proof of Theorem \ref{lem_NSP1}} 
\vspace*{-1mm}

In order to proof Theorem \ref{lem_NSP1}, we consider an arbitrary but fixed nLasso solution $\widehat{\vw} = \big( \big( \widehat{\vw}^{(1)} \big)^{T},\ldots,  \big( \widehat{\vw}^{(\numnodes)} \big)^{T} \big)^{T} $ 
(see \eqref{LNLprob}) and denote the estimation error between $\widehat{\vw}^{(\nodeidx)}$ 
and the true underlying weights $\overline{\vw}^{(\nodeidx)}$ (see \eqref{equ_lin_model}) as $\widetilde{\vw}^{(\nodeidx)} \defeq \widehat{\vw}^{(\nodeidx)} - \overline{\vw}^{(\nodeidx)}$. 

By the definition of nLasso \eqref{optProb},
\begin{equation}
\label{equ_inequ_basic_1}
\sum_{\nodeidx \in \samplingset} |\hat{y}^{(\nodeidx)}\!-\!y^{(\nodeidx)}| \!+\! \lambda \| \widehat{\vw} \|_{\rm TV}  \!\leq\! \sum_{\nodeidx \in \samplingset} |\noise^{(\nodeidx)}|_1 \!+\!  \lambda \| \overline{\vw} \|_{\rm TV}.\hspace{-1mm}
\end{equation} 
Since the true weight vectors $\overline{\vw}^{(\nodeidx)}$ are piece-wise constant (see \eqref{equ_def_clustered_signal_model}),  
$\| \overline{\vw} \|_{\compbound}=0$ and $\| \widetilde{\vw} \|_{\compbound}=\| \widehat{\vw} \|_{\compbound}$. 
Using the decomposition property and triangle inequality for the TV in \eqref{equ_inequ_basic_1}, 
\begin{align}
\sum_{\nodeidx\!\in\!\samplingset} |\hat{y}^{(\nodeidx)}\!-\!y^{(\nodeidx)}|\!+\!\lambda \| \widehat{\vw} \|_{\compbound}\!\leq\!\sum_{\nodeidx\in \samplingset} |\noise^{(\nodeidx)}| \!+\!\lambda \| \overline{\vw}\|_{\partial \partition}\!-\!\lambda \| \widehat{\vw} \|_{\partial \partition} \nonumber
 \end{align}
and, in turn, 
\vspace*{-1mm}
\begin{align}
\sum_{\nodeidx \in \samplingset} |\hat{y}^{(\nodeidx)} \!-\! y^{(\nodeidx)}|\!+\!\lambda \| \widetilde{\vw} \|_{\compbound} \leq \sum_{\nodeidx \in \samplingset} |\noise^{(\nodeidx)}| \!+\! \lambda \| \widetilde{\vw} \|_{\partial \partition}.
\vspace*{-1mm}
\label{equ_inequ_basic_2}
\end{align}
We conclude from \eqref{equ_inequ_basic_2} that
\vspace*{-1mm}
\begin{align}
\label{equ_upper_bound_complement_partition}
 \| \widetilde{\vw} \|_{\compbound}  &\leq (1/\lambda) \sum_{\nodeidx \in \samplingset} \big|\noise^{(\nodeidx)}\big| + \| \widetilde{\vw} \|_{\partial \partition},
\end{align}
Thus, for small noise $\noise^{(\nodeidx)}$ (see \eqref{equ_lin_model}), the nLasso estimation error $\widetilde{\vw}$ is piece-wise constant. 
However, it remains to control the size of the error for which we will invoke the NCC \ref{equ_ineq_multcompcondition_condition}. 

We can develop the LHS of \eqref{equ_inequ_basic_2} as 
\begin{align}
\sum_{\nodeidx \in \samplingset} |\hat{y}^{(\nodeidx)} \!-\!y^{(\nodeidx)} | &\stackrel{\eqref{equ_lin_model},\eqref{equ_predicted_label}}{=} 
\sum_{\nodeidx \in \samplingset} \big| \big( \vx^{(\nodeidx)}\big)^{T} \widetilde{\vw}^{(\nodeidx)} - \noise^{(\nodeidx)} \big| \nonumber\\
&\geq  \sum_{\nodeidx \in \samplingset}\big| \big( \vx^{(\nodeidx)}\big)^{T} \widetilde{\vw}^{(\nodeidx)}\big| \!-\! \sum_{\nodeidx \in \samplingset} |\noise^{(\nodeidx)}|, \label{equ_developed_LHS}
\end{align}
where we have used the triangle inequality in the last step. 
Combining \eqref{equ_developed_LHS} with \eqref{equ_inequ_basic_2}, 
\begin{align}
\hspace{-3mm} \sum_{\nodeidx \in \samplingset}\big| \big( \vx^{(\nodeidx)}\big)^{T} \widetilde{\vw}^{(\nodeidx)}\big|\!+\! \lambda \| \widetilde{\vw} \|_{\compbound}  \leq 2 \sum_{\nodeidx \in \samplingset} |\noise^{(\nodeidx)}| \!+\! \lambda \| \widetilde{\vw} \|_{\partial \partition}.\hspace{-1mm}
\label{equ_inequ_basic_3}
\end{align}
Since we assume NNC holds for $\samplingset$, \eqref{equ_ineq_multcompcondition_condition} yields
\begin{equation} 
\label{equ_inequ_diff_signal}
 (L/\sqrt{\sigdim}) \|\widetilde{\vw}  \|_{\partial \partition} \leq K  \sum_{\nodeidx \in \samplingset}\big| \big( \vx^{(\nodeidx)}\big)^{T} \widetilde{\vw}^{(\nodeidx)}\big| \!+\! \| \widetilde{\vw} \|_{\compbound}. 
\end{equation} 
Inserting \eqref{equ_inequ_diff_signal} into \eqref{equ_inequ_basic_3} and using $\lambda \defeq 1/K$, yields 
\begin{equation}
\label{equ_upper_bound_partial_partition}
\lambda(L/\sqrt{\sigdim}-1)\| \widetilde{\vw} \|_{\partial \partition} \leq 2  \sum_{\nodeidx \in \samplingset} |\noise^{(\nodeidx)} |.
\vspace*{-2mm}
\end{equation}  
Combining \eqref{equ_upper_bound_complement_partition} with \eqref{equ_upper_bound_partial_partition} yields  
\begin{align} 
\hspace*{-10mm}\| \widetilde{\vw} \|_{\rm TV}  &\!=\!\| \widetilde{\vw} \|_{\compbound}\!+\!\| \widetilde{\vw} \|_{\partial \partition} \nonumber\\
 &\stackrel{\eqref{equ_upper_bound_complement_partition}}{\leq}   \hspace*{-1mm} (1/\lambda) \sum_{\nodeidx \in \samplingset} |\noise^{(\nodeidx)}|_1 \!+\! 2 \| \widetilde{\vw} \|_{\partial \partition}  \nonumber \\
&  \stackrel{\eqref{equ_upper_bound_partial_partition}}{\leq}   \hspace*{-1mm} ( (1/\lambda) \!+\!\frac{4\sqrt{\sigdim}/\lambda}{(L\!-\!\sqrt{\sigdim})}) \hspace*{-1mm} \sum_{\nodeidx \in \samplingset}\hspace*{-1mm} |\noise^{(\nodeidx)} |.\nonumber 
\end{align}

\end{document}